\documentclass[11pt]{article}
\usepackage[final]{acl}

\usepackage{times}
\usepackage{latexsym}
\usepackage[T1]{fontenc}
\usepackage{enumitem}
\usepackage[utf8]{inputenc}
\usepackage{microtype}
\usepackage{graphicx}
\usepackage{booktabs}
\usepackage{amsmath}
\usepackage{amssymb}
\usepackage{multirow}
\usepackage{url}

\title{Cited but Unused: Parametric Provenance\\
       in Retrieval-Augmented Generation}

\author{
  Tanzeel Abbas$^{1}$ \quad Tehseen Zia$^{1}$ \quad David Windridge$^{2}$ \\[2pt]
  $^{1}$COMSATS University Islamabad \quad $^{2}$Middlesex University London \\[2pt]
  \texttt{tanzeelabbas114@gmail.com} \quad \texttt{tehseen.zia@comsats.edu.pk} \quad \texttt{d.windridge@mdx.ac.uk}
}

\begin{document}
\maketitle

\begin{abstract}

Retrieval-augmented generation systems attach citations to attribute claims to their sources. These citations frequently misstate provenance, and standard verification cannot detect this. We measure provenance by {\em counterfactual value substitution}: a fact in the retrieved passages is replaced with a false variant and the model re-queried. We find that across five sizes of one model
family and a second comparator family, 1.2\% to 15.7\% of citable claims derive from parametric
memory rather than the retrieved context. At the most conservative substitution distance these figures are 4.6\% and 8.5\% on the two models we tested at it, with the rate rising with model size within a family. We further find that natural language inference verification, the standard approach, gives results indistinguishable from chance across models, even when restricted to the cited passage. We determine that the cause is structural: a model answering from memory selects a citation that agrees with its output.
Comparing ten attribution methods on identical claims, the contrast between the model's distribution with and without context best separates provenance. Acting on that signal does not improve provenance honesty at these base rates. The resulting detector is hence still not accurate enough to act on: it measures what the model {\em knew} rather
than what it {\em used}, and separating those remains the open problem.
\end{abstract}

\section{Introduction}

Retrieval-Augmented Generation (RAG) systems present generated claims alongside citations
to retrieved passages. The citation identifies the evidence required to verify the claim. Its utility depends on being accurate about provenance, not merely about support: a passage that agrees with a claim it did not produce directs the reader to evidence lacking confirmatory power.

Verification typically asks whether a cited passage supports the claim \citep{rashkin2023measuring,gao2023enabling}, though support is necessary rather than sufficient. Selecting citations after an answer is written produces agreement by construction, so the check passes even when the citation misstates its origin.
\citet{wallat2025correctness} terms this tendency {\em post-rationalisation}.

Measuring this requires ground truth that does not rest on inferences about internal states. We therefore propose {\em counterfactual value substitution}: a value present in both the answer and the retrieved passages is replaced throughout those passages, titles included, with a false variant, and the question asked again. A model reporting the false value depended on the context; one repeating the original used its memory. The label is therefore constructed rather than inferred, which permits measurement at scale.

In empirically testing this approach, we find that provenance failure affects 1.2\% to 15.7\% of citable claims depending on the model, at our default substitution distance. This
rate rises with size within a model family, and differs twofold between families of
comparable size. We also report an artefact of the proposed process: our default substitution approach edited passage
bodies but not passage titles; correcting for this roughly halved the measured rates.

Our contributions are as follows:

\begin{enumerate}[noitemsep,topsep=2pt,leftmargin=*]
\item A measurement of parametric provenance across five sizes of one model family
and a second family, separating the effect of scale from that of family
(Section~\ref{sec:rates}).
\item A concrete determination of characteristic citation behaviour: claims answered from memory carry
citations just as readily as claims answered from the passages
(Section~\ref{sec:citations}).
\item Evidence that support-based verification is structurally blind to this
failure, being indistinguishable from chance on every model (Section~\ref{sec:nli}).
\item A comparison of ten attribution methods on identical claims, showing they
group by what they measure ({\em token contrast, sentence contrast, passage attribution, support/preference}) and that passage-level attribution separates provenance
only weakly (Section~\ref{sec:comparison}).
\item A localisation result: the provenance signal is carried by the
model's distribution \emph{before} context is supplied, and measuring at the
disputed token rather than over the sentence improves detection
(Section~\ref{sec:localised}).
\item An end-to-end evaluation showing that detection does not yield deployable flagging at these base rates (Section~\ref{sec:repair}).
\item A measurement artefact in entity-substitution protocols that inflates measured rates roughly twofold, with its correction (Section~\ref{sec:artefact}). We also find that the largest model detects the manipulation and says so (Section~\ref{sec:conflict}).
\end{enumerate}

\section{Related Work}

\paragraph{Retrieval-augmented generation and attribution.}
Retrieval augmentation \citep{lewis2020retrieval} grounds generation in retrieved passages, typically obtained by dense retrieval \citep{karpukhin2020dense,ni2022large}. Attribution was formalised by \citet{rashkin2023measuring} as {\em attributable-to-identified-sources}; \citet{bohnet2022attributed} introduced AutoAIS as an NLI proxy, and \citet{gao2023enabling} proposed ALCE, whose citation protocol has become the {\em de facto} standard. \citet{liu2023evaluating} found roughly half of sentences in generative search engines fully supported, \citet{yue2023automatic} and \citet{min2023factscore} refined automatic scoring, and \citet{niu2024ragtruth} evaluated span-level annotations. Each measures the degree to which a passage \emph{supports} a claim.

\paragraph{Improving citations.} A large body of work seeks to improve citation quality, under the corroborative definition, by revising statements to match evidence \citep{gao2023rarr}, reassigning citations \citep{maheshwari2025citefix,li2024citation}, or training models to be better-supported \citep{asai2024selfrag,ye2024effective,slobodkin2024attribute,song2025measuring}. However, all optimise a criterion that we will argue in Section~\ref{sec:nli} is uninformative in regard to provenance.

\paragraph{Correctness versus faithfulness.} The distinction we adopt derives from \citet{jacovi2020towards} on faithfulness against plausibility, and from evidence that models {\em rationalise after the fact} \citep{turpin2023language}. \citet{wallat2025correctness} brought this to citation, measuring post-rationalisation on one large commercial model.
\citet{vandort2026howcite} gave a mechanistic account of post-rationalisation on one 8B model, finding that citation generation driven by shallow heuristics runs in parallel to answer generation. \citet{worledge2024unifying} supply the corroborative and contributive
terminology that we shall adopt. Our contribution is both quantitative and methodological: multiple models, and a protocol whose labels are {\em constructed} rather than {\em inferred}.

\paragraph{Context attribution.} Context attribution measures the influence of parts of the input on the output. ContextCite \citep{cohenwang2024contextcite} fits a sparse surrogate over randomised ablations; MIRAGE \citep{qi2024mirage} and PECoRe \citep{sarti2024quantifying} use internals; ARC-JSD \citep{li2026arcjsd} uses divergence under sentence ablation; SelfCite \citep{chuang2025selfcite} defines necessity and sufficiency rewards; AttriBoT \citep{liu2025attribot} makes leave-one-out efficient. All are deletion-based. We will seek to demonstrate in Section~\ref{sec:deletion} that deletion-based margins inherently separate provenance only weakly.

\paragraph{Parametric versus contextual knowledge.} \citet{longpre2021entity}
introduced entity substitution in the context of question answering, which will form the basis of our protocol and the associated artefact of Section~\ref{sec:artefact}. \citet{mallen2023when} showed that parametric memory suffices for popular entities but fails for rare ones;
\citet{xie2024adaptive} and \citet{neeman2023disentqa} studied how models arbitrate
between memory and counter-evidence; \citet{wu2024clasheval}, \citet{su2024conflictbank} and
\citet{ming2025faitheval} provide benchmarks, and \citet{xu2024knowledge} survey the
area as a whole. A parallel line studies conflict among retrieved documents rather than between memory and context \citep{yuan2026benchmarking}, asking which source should prevail; we ask which source an answer came from.

Methods for controlling the balance between memory and context include context-faithful prompting
\citep{zhou2023context} and context-aware decoding \citep{shi2024trusting}.
Other quantitative work locates the competition in specific components
\citep{monea2024glitch, wu2025retrieval}. Closest to our approach, ConSens
\citep{consens2025} contrasts with- and without-context perplexity while TrustMargin
\citep{xu2026trustmargin} arbitrates using likelihood margins; both seek score quantification over whole answers, as does the consistency checking approach of \citet{manakul2023selfcheckgpt}. We will show that localisation to the disputed token does indeed matter, and that these measures computed with the context present are uninformative.

\section{Measuring Provenance}

Let $q$ be a question, $C = (d_1, \dots, d_k)$ the retrieved passages, and $c$ a
generated claim carrying a citation to $d_j$. We write $Z \in \{\textrm{contextual},
\textrm{parametric}\}$ for the latent provenance of $c$.

Following \citet{wallat2025correctness}, a citation is \textbf{correct} if $d_j$
entails $c$, and \textbf{faithful} if it is correct \emph{and} $c$ was causally
influenced by $d_j$. A post-rationalised citation is correct but not faithful; this
is exactly the circumstance that support-based verification cannot detect, since the property it tests is already satisfied.

$Z$ itself is not directly observable, and so we define it operationally through intervention: a claim is parametric if {\em the model reports the same value when the passages are edited to assert a different value}. This definition is both behavioural and testable by construction, which allows labelling without human judgement of intent.

\paragraph{Protocol.} We locate a value $v$ appearing in both the answer and the
retrieved passages, replacing it throughout the passages \emph{including their titles} with a false variant $v'$, and then regenerate. If the model reports $v'$ then it used the context; if it reports $v$ it instead relied on memorisation; if it reports neither, or both, the case is discarded as being undetermined. Cases in which $v$ appears in the question are excluded, since the question cannot be edited and its provenance is undecidable.
Both of these design choices follow from the artefact analysis of
Section~\ref{sec:artefact}.

\section{Experimental Setup}

We investigate five sizes of Qwen2.5-Instruct, at 0.5B, 1.5B, 3B, 7B and 14B parameters
\citep{qwen2025qwen25}, and Llama-3.1-8B-Instruct \citep{grattafiori2024llama}, all
in bfloat16. The Qwen series hence spans a thirtyfold range in parameter count within one
family, allowing separation of the effect of scale from the effect of family. Prompts
are the ALCE defaults \citep{gao2023enabling} with two demonstrations in completion
style. Instruction-tuned models given a chat template largely ignore the citation format,
and so all generation is completion-style.

Our primary reference dataset is ASQA \citep{stelmakh2022asqa} with GTR \citep{ni2022large}
top-5 retrieval, 800 questions per condition. We also assess ELI5
\citep{fan2019eli5} with BM25 retrieval and QAMPARI \citep{amouyal2023qampari} using
Qwen2.5-7B, to establish whether the provenance misattribution phenomenon is specific to factoid questions and dense retrieval. Decoding is greedy, and hence generation is deterministic given the model and prompt; the single stochastic element is the choice of counterfactual value, which is here seeded and recorded for individual runs. We do not repeat runs under differing seeds, and so the intervals reported reflect sampling over claims rather than over counterfactual value choices.

The Qwen2.5 series is released under Apache 2.0 and Llama-3.1 under the Llama 3.1
Community License, both permitting research use. ASQA, ELI5 and QAMPARI are public
benchmarks derived from Wikipedia and web text, used here for their intended purpose
of evaluation.

The support-based baseline is a DeBERTa \citep{he2021deberta} cross-encoder trained
for NLI. We apply it in its strongest form, taking for each claim the maximum
entailment over all retrieved passages. The baseline is therefore not penalised for
citation errors. Confidence intervals for rates are Wilson intervals
\citep{wilson1927probable}. AUROC intervals and paired comparisons use a bootstrap
over claims with 2000 resamples \citep{efron1993introduction}. Resampling claims
rather than scores preserves the correlation between detector pairs evaluated on identical items, which makes the comparison more direct than one based on marginal
intervals. Annotator agreement is reported as Cohen's $\kappa$ \citep{cohen1960coefficient}.

\section{Prevalence}
\label{sec:rates}

\begin{table}[t]
\centering\small
\setlength{\tabcolsep}{4pt}
\begin{tabular}{lrrrr}
\toprule
Model & B & Rate & 95\% CI & par./dec. \\
\midrule
Qwen2.5-0.5B & 0.5 & 1.2 & [0.6, 2.6] & 6 / 498 \\
Qwen2.5-1.5B & 1.5 & 6.5 & [4.5, 9.3] & 27 / 417 \\
Qwen2.5-3B & 3.0 & 8.3 & [6.1, 11.2] & 38 / 456 \\
Qwen2.5-7B & 7.6 & 8.2 & [6.0, 11.2] & 35 / 426 \\
Qwen2.5-14B & 14.8 & 13.3 & [10.4, 16.9] & 56 / 420 \\
Llama-3.1-8B & 8.0 & 15.7 & [12.6, 19.4] & 70 / 446 \\
\bottomrule
\end{tabular}
\caption{Parametric provenance rate. An answer counts as contextual whenever it
asserts the substituted value, including cases where the model reports this value
while asserting a source mismatch.}
\label{tab:rates}
\end{table}

\begin{figure}[t]
\centering
\includegraphics[width=\columnwidth]{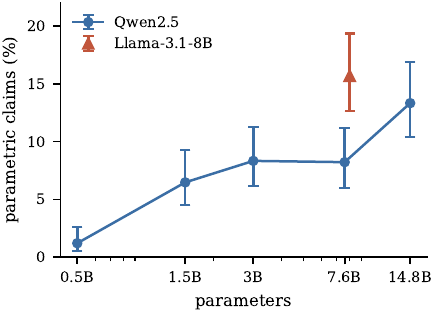}
\caption{Parametric provenance rate against parameter count. Bars are Wilson 95\% intervals.
Llama-3.1-8B sits above every Qwen model including one twice its size, indicating that family
matters independently of scale.}
\label{fig:scaling}
\end{figure}

Table~\ref{tab:rates} and Figure~\ref{fig:scaling} report rates on ASQA. Within the Qwen family the rate rises nearly monotonically across a thirtyfold parameter range, from 1.2\% at 0.5B to 13.3\% at 14B, with Kendall's $\tau = 0.80$. The rise holds within one family trained on the same data, so it is not an artefact of cross-family comparison. The evidence is driven by the endpoints: five of ten pairwise comparisons have non-overlapping intervals, with 0.5B and 1.5B falling clearly below 14B and the trend between 3B, 7B and 14B less distinct.

Family is at least as significant as size: Llama-3.1-8B reaches 15.7\% against 8.2\% for Qwen2.5-7B at almost the same parameter count, higher than any Qwen model including one twice its size.

For ELI5 and QAMPARI evaluations, the respective rates are 4.5\% (4 of 89 decided) and 2.4\% (3 of 124) on the reference Qwen2.5-7B model. The misattribution phenomenon is hence apparent in both evaluations, but overall coverage is far lower; only, respectively, 132 and 174 of 800
answers contain an editable value, as against 542 to 613 for ASQA. Long-form explanations and entity lists contain fewer dated facts. Labels also arise from string containment, which can match a value inside a larger token: one of the 35 parametric instances matched ``1982'' inside ``1982-83''. Restricting to values of
three or more digits moves the associated rate from 8.2\% to 8.0\%. With so few positive cases we hence report these rates here for completeness, basing our principal detection claims on ASQA.

\begin{table}[t]
\centering\small
\setlength{\tabcolsep}{4pt}
\begin{tabular}{lrcrc}
\toprule
& \multicolumn{2}{c}{Qwen2.5-7B} & \multicolumn{2}{c}{Qwen2.5-14B} \\
\cmidrule(lr){2-3}\cmidrule(lr){4-5}
Substitution & Rate & 95\% CI & Rate & 95\% CI \\
\midrule
Near & 4.6 & [3.0, 6.8] & 8.5 & [6.3, 11.3] \\
Default & 8.2 & [6.0, 11.2] & 13.3 & [10.4, 16.9] \\
Far & 16.3 & [12.9, 20.3] & 17.6 & [14.2, 21.5] \\
\bottomrule
\end{tabular}
\caption{Parametric provenance rate against the extent to which the counterfactual value {\em date} is
perturbed: 2 to 5 years (near), 18 to 40 (default), or 120 to 200 (far).}
\label{tab:ablation}
\end{table}

\paragraph{Robustness.} Table~\ref{tab:ablation} varies how far the counterfactual date is perturbed. The rate rises at every increment on both models, so the figures in Table~\ref{tab:rates} are conditional on that choice. The rate for Qwen2.5-14B is higher than Qwen2.5-7B at all three settings, but the intervals overlap in each case. The gap narrows from 1.85 to 1.08 as the substitution becomes more extreme. The ablation study in Appendix~\ref{sec:ablations}
adds two further parameters. Retrieval depth does not affect the above findings, with the rate
statistically unchanged from a depth of one to a depth of ten passages ($p = 0.47$). However, demonstration
count does appear to be significant, with parametric provenance rates rising from 5.4\% to 12.6\% between counts of zero and four ($p = 0.0003$).

\subsection{Larger models are able to better detect counterfactual manipulation}
\label{sec:conflict}

The protocol has two outcomes by design; the models produce a third, reporting the value the passages assert, stating the value held in weights, and objecting that the source appears wrong (Appendix~\ref{sec:examples}).

The model has neither deferred to the passage nor answered from memory: it has detected the manipulation. Because both values appear the string-matching rule returns no decision, so these cases were discarded as undetermined. We now consider them explicitly.

We hence label a case \emph{conflict-flagged} when both values appear \emph{and} the
answer objects to the source. Both conditions are required: reporting the two values
without objecting represents a state of ambiguity; objecting without stating the true value reveals nothing about what the model knows.

\begin{table}[t]
\centering\small
\begin{tabular}{lrr}
\toprule
Model & Cases & Share of records \\
\midrule
Qwen2.5-0.5B & 0 & 0.0 \\
Qwen2.5-1.5B & 0 & 0.0 \\
Qwen2.5-3B & 0 & 0.0 \\
Qwen2.5-7B & 1 & 0.2 \\
Qwen2.5-14B & 140 & 25.7 \\
Llama-3.1-8B & 5 & 0.9 \\
\bottomrule
\end{tabular}
\caption{Conflict-flagged cases: the model reports the retrieved value, states the
value it holds, and objects to the source.}
\label{tab:conflict}
\end{table}

Table~\ref{tab:conflict} gives the counts for this amended protocol. The indicated behaviour is absent below 7B, appears
once at 7B, and accounts for a quarter of Qwen2.5-14B's records. This is hence a step threshold
rather than a gradient over the measured range.

Two further checks support this reading, as set out in Appendix~\ref{sec:examples}.

\subsection{Reliability of labels}
\label{sec:validation}

An earlier blind round, conducted before the protocol was amended, disagreed with the automatic labels on 18 of 97 decidable cases. Fifteen of those disagreements shared a single cause, which Section~\ref{sec:artefact} reports. The experimental round described below hence incorporates this correction.

Three annotators (none an author) each judged an identical set of 120 items drawn from
Qwen2.5-14B and balanced between the classes. Each item indicated the question, the passages as the model received them, and the answer given, with the disputed value unmarked and the automatic label withheld.

Human agreement with the automated labels was, respectively, 81.7\%, 78.0\% and 76.9\%
($\kappa = 0.637$, 0.568 and 0.537), and 82.8\% for the majority vote
($\kappa = 0.658$) over the 116 items having a label. Pairwise agreement between the
annotators themselves, a measure of inter-rater reliability, is lower, being 74.8\% to 81.4\% ($\kappa$ 0.486 to 0.604). The labels therefore sit closer to the consensus than does any single
annotator, which is perhaps the more useful comparison for a protocol intended to
replace human judgement at scale.

The disagreements are directional. Of the 20 items where the majority differs from
the label, 18 are cases where the protocol indicates {\em contextual} and the annotators indicate
{\em parametric}, with 15 of these following the passage while objecting to it. An answer that is plainly false is read as model error even when it reproduces the passage faithfully. This is a limitation of blind validation on this task rather than evidence against the labels.

\subsection{Parametric claims are cited}
\label{sec:citations}

It matters whether claims answered from memory are cited at all. Of the four models above 1.5B, between 89\% and 100\% of parametric claims carry a citation, in each case within a few points of that model's contextual rate (Table~\ref{tab:citations}). A model answering from memory hence cites as readily as one answering
from the passages.

\begin{table}[t]
\centering\footnotesize
\setlength{\tabcolsep}{3pt}
\begin{tabular}{lrrrr}
\toprule
& \multicolumn{2}{c}{Param.} & \multicolumn{2}{c}{Context.} \\
\cmidrule(lr){2-3}\cmidrule(lr){4-5}
Model & cited & \% & cited & \% \\
\midrule
Qwen2.5-0.5B & 1 / 6 & 17 & 58 / 491 & 12 \\
Qwen2.5-1.5B & 12 / 27 & 44 & 206 / 388 & 53 \\
Qwen2.5-3B & 36 / 38 & 95 & 392 / 417 & 94 \\
Qwen2.5-7B & 35 / 35 & 100 & 373 / 390 & 96 \\
Qwen2.5-14B & 50 / 56 & 89 & 209 / 224 & 93 \\
Llama-3.1-8B & 66 / 70 & 94 & 362 / 371 & 98 \\
\bottomrule
\end{tabular}
\caption{Claims carrying at least one citation, by provenance. A claim is cited if
the sentence asserting the disputed value contains a citation marker.}
\label{tab:citations}
\end{table}

The two smallest models cite far less often in both classes, reflecting weaker adherence to the format rather than anything about provenance. That these citations cannot support the claims they carry follows from the protocol rather than from measurement. The substitution removes the asserted value from every
passage containing it, and so a parametric claim necessarily cites a passage asserting
something different. Contextual claims cite a passage containing their value in 85.7\% to 99.5\% of cases
on those same four models. These models cite correctly when the answer derives from the
context, and so the parametric cases are a failure of provenance rather than of citation
ability.

\section{Detection}

We ask whether the failure can be identified without regenerating the answer, evaluating two approaches from the literature and then the model's own distribution.

\subsection{Support-based verification is at the chance level}
\label{sec:nli}

We apply a DeBERTa NLI verifier in its strongest form:
for each claim we take the maximum entailment over all retrieved passages. The \texttt{nli\_entailment} row of Table~\ref{tab:comparison} gives the result:
every single bootstrap interval contains the chance value of 0.5. Appendix~\ref{sec:nlisweep} repeats this assessment
with three further entailment models, with the same outcome. On Qwen2.5-1.5B parametric claims are \emph{better} entailed, and so a support-based
checker would rank them as more trustworthy. These findings follow from the mechanism itself, and not from a weak verifier: a model answering from memory selects a citation agreeing with what it has written.

Restricting the check to the cited passage rather than the best passage does not
help: on Qwen2.5-7B it gives AUROC 0.532, interval [0.415, 0.646], over the 408
claims carrying a citation. This is the most direct test that can be run. Our substitution removes the asserted value from every passage, so the cited passage contains the value a parametric claim asserts in 0 of 35 cases, against 96.5\% of cited contextual claims. The contradiction sits in the text the verifier reads and it does not register: mean entailment is higher for parametric claims, 0.329 against 0.214.

\subsection{Attribution methods compared}
\label{sec:comparison}

\begin{table*}[t]
\centering\small
\setlength{\tabcolsep}{5pt}
\begin{tabular}{lrrrrrrr}
\toprule
Method & 0.5B & 1.5B & 3B & 7B & 14B & Llama & Mean \\
\midrule
Localised (ours) & 0.877 & 0.833 & 0.802 & 0.779 & 0.743 & 0.697 & 0.788 \\
JSD, no context & 0.900 & 0.844 & 0.820 & 0.756 & 0.744 & 0.671 & 0.789 \\
Sentence-level & 0.698 & 0.725 & 0.669 & 0.596 & 0.683 & 0.643 & 0.669 \\
ConSens & 0.717 & 0.751 & 0.701 & 0.619 & 0.739 & 0.665 & 0.699 \\
Sufficiency & 0.807 & 0.684 & 0.643 & 0.596 & 0.670 & 0.644 & 0.674 \\
LOO max & 0.667 & 0.640 & 0.557 & 0.541 & 0.656 & 0.616 & 0.613 \\
LOO sum & 0.653 & 0.637 & 0.552 & 0.559 & 0.679 & 0.611 & 0.615 \\
Necessity & 0.692 & 0.603 & 0.521 & 0.565 & 0.585 & 0.598 & 0.594 \\
NLI entailment & 0.553 & 0.452 & 0.505 & 0.526 & 0.506 & 0.540 & 0.514 \\
TrustMargin & 0.525 & 0.498 & 0.380 & 0.512 & 0.421 & 0.435 & 0.462 \\
\bottomrule
\end{tabular}
\caption{Attribution methods scored on identical claims and labels, AUROC. None was
designed to identify provenance: they ask whether a passage supports a claim, or how
much it contributes to the claim's likelihood. The table therefore tests whether
contribution and origin coincide. Methods group by what they measure rather than by
lineage: the token-level context contrast (top two rows) separates provenance best of all,
the sentence-level contrast is next best with passage-level attribution below that, and
support-based checking not at all.}
\label{tab:comparison}
\end{table*}

\begin{figure}[t]
\centering
\includegraphics[width=\columnwidth]{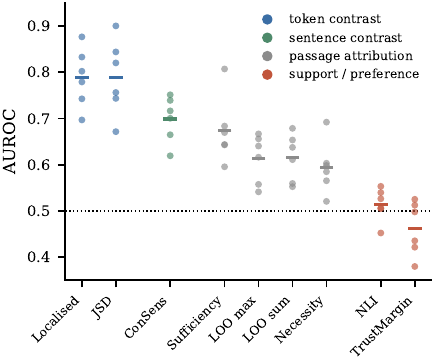}
\caption{Attribution methods grouped in terms of what they measure. Each plot point is one model;
the bar is the group mean. Methods hence cluster by quantity rather than by lineage.}
\label{fig:methods}
\end{figure}

Figure~\ref{fig:methods} suggests that the methods fall into four groups based on what they measure, with the groups ordering
cleanly. Measuring the context contrast at the disputed token gives a mean AUROC of 0.79.
Measuring the same contrast over the sentence gives 0.70 for ConSens and 0.67 for
the sentence-level likelihood, and so resolution costs roughly 0.12 AUROC. Passage-level
attribution, which is what leave-one-out and the necessity and sufficiency rewards
compute, gives 0.59 to 0.67. Resolution
matters more than method: ConSens and our signal ask the same question at different
granularity, and the finer granularity is worth 0.09 AUROC.

Our signal is statistically indistinguishable from the token-level Jensen-Shannon
divergence. Both compare the distribution with and without context at the same
position. We take this as a finding about which quantity best separates provenance, not
as evidence for a new method. Against the support-based baseline, necessity and TrustMargin, the paired difference
is significant on all six models, and against leave-one-out on four of the models.

\subsection{Deletion-based attribution separates provenance weakly}
\label{sec:deletion}

The dominant approach to context attribution measures the likelihood drop when a
passage is removed \citep{cohenwang2024contextcite,li2026arcjsd,chuang2025selfcite}.
Applied to provenance, leave-one-out reaches AUROC 0.541 to 0.667 across these six
models, and removing the whole context reaches AUROC 0.619 to 0.751. That is above chance
but well below the token-level contrast, and it is the weakest of the
likelihood-based families in Table~\ref{tab:comparison}.

We also run ContextCite itself, whose sparse surrogate is fitted over 64 randomised
ablations per claim rather than a single leave-one-out pass. On Qwen2.5-7B it reaches AUROC 0.507, with a bootstrap
interval of [0.413, 0.605] that contains the chance value 0.5. That is slightly below our
leave-one-out figure of 0.541 on the same claims.

In 32 instances the computation extracts no further signal, the reason being, we believe, that deletion effectively answers a different question. Removing a passage
withdraws support for every individual candidate value at once rather than forcing resolution between
them. Where the evidence is redundant the model can still produce the same value from
its remaining sources or from memory, and so the performance drop is small regardless of which
source drove the claim. Deletion thus measures how much a passage contributes to a
claim's likelihood, which is distinct from the question of whether the claim originated from it.

\subsection{Localisation improves detection}
\label{sec:localised}

A sentence-level likelihood averages over roughly a dozen tokens, of which one
carries the disputed value. If the informative fraction is small, averaging costs
signal. We therefore measure the same quantities at the value token itself: its
log-probability, the entropy of the distribution at that position, and how each changes
when the context is removed. Table~\ref{tab:signals} reports the result.
Localisation improves detection on every model, and significantly on three of the six:
Qwen2.5-1.5B, 3B and 7B. On the smallest model the gain is the largest observed
($+0.178$) but rests on six positive cases and does not reach significance, and on
Qwen2.5-14B and Llama-3.1-8B the two signals are statistically indistinguishable. We
therefore do not claim a uniform improvement. Against the support-based baseline the
improvement is significant on every model.
ELI5 and QAMPARI yield only four and three positive cases, too few for inference; we
report their rates descriptively and draw no conclusions from significance tests on
them.

\subsection{Provenance information precedes the context}
\label{sec:precedes}

\begin{table}[t]
\centering\footnotesize
\setlength{\tabcolsep}{1.5pt}
\begin{tabular}{@{}lrrrrrr@{}}
\toprule
Signal & 0.5B & 1.5B & 3B & 7B & 14B & L-8B \\
\midrule
log p(v), no context & 0.678 & 0.811 & 0.811 & 0.731 & 0.733 & 0.673 \\
entropy, no context & 0.607 & 0.833 & 0.818 & 0.778 & 0.736 & 0.683 \\
change in log p & 0.855 & 0.816 & 0.805 & 0.721 & 0.748 & 0.695 \\
change in entropy & 0.877 & 0.833 & 0.802 & 0.776 & 0.743 & 0.697 \\
\textbf{log p(v), with context} & \textbf{0.191} & \textbf{0.475} & \textbf{0.576} & \textbf{0.502} & \textbf{0.477} & \textbf{0.457} \\
\textbf{entropy, with context} & \textbf{0.191} & \textbf{0.480} & \textbf{0.578} & \textbf{0.501} & \textbf{0.478} & \textbf{0.457} \\
\bottomrule
\end{tabular}
\caption{AUROC of the log-probability and entropy signals at the disputed token.
Quantities computed with the context present carry no usable provenance
information: their bootstrap intervals contain 0.5 on five of six models. On
Qwen2.5-0.5B the interval lies below chance, which we do not interpret given six
positive cases. Rank and margin-to-top variants of the same contrast behave alike,
averaging 0.68 to 0.71 without context, and are omitted for space. Values here may
differ from Table~\ref{tab:comparison} in the third decimal, since the two analyses
were run in different environments (Appendix~\ref{sec:repro}).}
\label{tab:signals}
\end{table}

Table~\ref{tab:signals} gives what we regard as the most
robust finding in this work. Quantities measured with the context present carry no usable provenance information. On five of the six models every bootstrap interval for these two signals contains the chance value 0.5, and on Qwen2.5-0.5B it lies below chance. All of the significant provenance information hence lies in the
distribution before the context is identified, and in how much the context changes this.

The mechanism explains this. A memorised claim is one the model is already confident about, so a supporting passage tells it little. Once that passage is in the prompt, a memorised claim and a context-grounded one look alike. A detector reading the model's
state at generation time, when the context is necessarily present, is therefore reading the
wrong quantity.

Asked with no context at all, the model produces the value anyway for 33\% to 84\% of
parametric claims against 10\% to 60\% of contextual ones, on every model.

Appendix~\ref{sec:reconcile} reconciles our rates with those previously reported,
Appendix~\ref{sec:examples} gives worked examples, and Appendix~\ref{sec:discarded}
reports three signals that did not separate the classes.

\paragraph{Scoring without knowing where to look.} The position scored above is one
our protocol chose, which a deployed detector would not know. Scoring every numeric
span of the answer instead costs nothing extra, since one forward pass yields logits
at every position. On Qwen2.5-7B this lowers AUROC from 0.760 to 0.668 (paired
bootstrap, $p = 0.002$), an interval that still excludes chance, so the signal
survives the loss of the oracle. Appendix~\ref{sec:candidates} gives the procedure,
and Table~\ref{tab:signals} should be read as an upper bound on a deployed detector.

\subsection{The signal measures knowledge, not use}
\label{sec:errors}

Partitioning claims by whether the model could produce the value with no context
indicates where detection fails. Among the 220 Qwen2.5-7B claims where it could,
13.2\% were parametric; among the 205 where it could not, 2.9\% were. Prior knowledge
raises the chance more than fourfold, and 29 of the 35 parametric claims fall in the
first group.

Knowing is not deciding: of those 220 claims, 191 followed the edited passage anyway.
Restricted to claims the model already knew, AUROC falls to 0.651; restricted to those
it did not, it reaches 0.812, against 0.776 overall. The signal measures what the model
knows rather than what it used, and the two coincide only where the context contradicts
that knowledge, which in ordinary retrieval it rarely does. We take this to be the main
obstacle to improving detection within this family of methods.

\section{Utility of Detection}
\label{sec:repair}

Define \textbf{provenance honesty} as the fraction of claims presented truthfully
in regard to their origin. Hence, a system that cites everything is honest exactly for its
contextual claims, so honesty without repair is $1 - \pi$ where $\pi$ is the
parametric rate.

Flagging trades one error for another. We write $c$ for the cost of missing a
parametric claim relative to the cost of a false flag. The expected gain at budget
$b$ is then $(\textrm{precision} \cdot (1+c) - 1) \cdot b$. Flagging is therefore
worthwhile only where
\[
\textrm{precision} > \frac{1}{1+c}.
\]

\begin{table}[t]
\centering\small
\begin{tabular}{lrrr}
\toprule
Model & Honesty & Gain & Cost red.\ 3:1 \\
\midrule
Qwen2.5-0.5B & 0.988 & $+0.000$ & 0.0 \\
Qwen2.5-1.5B & 0.935 & $+0.000$ & 7.4 \\
Qwen2.5-3B & 0.917 & $+0.000$ & 0.0 \\
Qwen2.5-7B & 0.918 & $-0.000$ & 0.0 \\
Qwen2.5-14B & 0.867 & $+0.000$ & 32.1 \\
Llama-3.1-8B & 0.843 & $+0.000$ & 2.9 \\
\bottomrule
\end{tabular}
\caption{End-to-end repair under symmetric costs. \emph{Honesty} is one minus the
rate in Table~\ref{tab:rates}; \emph{Gain} is the best improvement over all budgets.
The last column is the cost reduction at 3:1, over claims with detector scores.}
\label{tab:repair}
\end{table}

Table~\ref{tab:repair} indicates that flagging does not pay on any of the six models:
honesty falls at every budget. The single positive value is a gain of $+0.022$ on
ELI5, resting on three correct flags and one incorrect, which we regard as noise. At base rates between 1.2\% and 15.7\%, precision peaks at 0.45, below the
break-even point of 0.5, so every flag costs more honesty than it buys.
Appendix~\ref{sec:precision} shows the curves. The two errors are not equally harmful: a missed parametric claim asserts false
provenance, whereas a false flag merely asks the reader to check a sound citation.
Under asymmetric costs the largest gain is a 46\% cost reduction on Qwen2.5-14B, and
only at a 5:1 ratio and a 35\% budget.

We regard this as a useful negative: detection quality and deployability are
different properties, and the gap is governed by precision at the operating point
rather than by ranking quality.

\section{A Measurement Artefact in\\Entity-Substitution Protocols}
\label{sec:artefact}

Fifteen of the eighteen disagreements in our first blind round
(Section~\ref{sec:validation}) shared a single cause. Our substitution replaced the target value in passage bodies
but not in passage \emph{titles}. A passage titled ``2016 Summer Olympics'' whose body
was edited to read ``2029'' lets the model read the original value from the heading, a
contextual act our protocol labels parametric.

Of the cases originally labelled parametric, between 31\% and 95\% had the true value
still visible either in a title or in the question. The corrected protocol substitutes the
value in titles as well as bodies, and discards any example where it survives in the
question. Table~\ref{tab:artefact} gives the effect.

Any protocol built on entity substitution is exposed to this, and our own uncorrected
numbers overstated the phenomenon roughly twofold.

\section{Conclusion}

A citation carries two claims: that the passage supports the statement, and that the
passage produced it. Evaluation measures the first. We measure the second and find it
fails for 1.2\% to 15.7\% of citable claims. The rate rises with model size within a
family, but family matters at least as much: at comparable size our two families
differ twofold.

Support-based checking cannot detect this, because a post-rationalised citation does
entail the claim it carries. Across ten attribution methods, what separates provenance
is the context contrast at the disputed token, before that context arrives.

The detector is not accurate enough to act on: it measures what the model {\em knew}
rather than what it {\em used}, and separating those is the open problem. 

\section*{Limitations}

\paragraph{Model scale.} We evaluate open-weight models from 0.5B to 14B
parameters. Our method requires token-level log-probabilities with and without the retrieved
context, which closed frontier models do not expose. The trend within the Qwen family suggests larger models
would be more affected, but the largest we measure is 14B and the extrapolation is
not tested.

\paragraph{Undetermined cases.} Between 19\% and 24\% of records are discarded as
undetermined: the answer asserted neither the true nor the substituted value. If
those cases differ systematically from the decided ones, the rates are biased. The
protocol provides no label for them, so we cannot check this from within it.

\paragraph{Coverage.} The protocol requires a numeric value shared between answer
and passages: roughly 70\% of ASQA answers but only 16.5\% and 21.8\% of ELI5
and QAMPARI. Claims without such a value are not measured, and we cannot say whether
they behave alike.

\paragraph{Sample size.} ELI5 and QAMPARI yield four and three parametric cases,
too few for detection statistics. Our detection claims rest on ASQA. The one
positive honesty gain in our corrected results comes from ELI5 and rests on a
handful of claims, as noted in Section~\ref{sec:repair}.

\paragraph{Operational definition.} We define provenance behaviourally, through
response to a counterfactual edit. A model might use a passage yet still resist an
edit, which our protocol would label parametric. Our annotators' disagreements run the other way. Of their 20 differences from the
automatic labels, 18 are cases the protocol calls contextual and they call
parametric, so by their reading it undercounts rather than overcounts. The label therefore reflects the
provenance of the disputed value rather than of the whole claim.

\paragraph{Validation scope.} The blind round covers 120 items from one model,
Qwen2.5-14B. Whether the labels are equally reliable on the other five is untested,
and $\kappa = 0.658$ for the majority is substantial rather than strong.

\paragraph{Baselines we implement rather than run.} Our comparison table uses reimplementations of the quantity each method computes
rather than the published systems. The exception is ContextCite, which we run as
published on Qwen2.5-7B (\S\ref{sec:deletion}). Our claim about generation-time detection rests on
output-distribution signals only: methods that read attention or hidden states
\citep{qi2024mirage,wu2025retrieval} are cited but not run, and could carry
information the output distribution does not.

\paragraph{Value handling.} Three weaknesses share a common cause: the target value is
found and replaced by string operations rather than by parsing. Matching can catch a
value inside a longer token, whose size Section~\ref{sec:rates} reports. Extraction
can split a comma-separated number, taking 000 from 4,000. And the substitution
treats a value as a year only within 1600 to 2029, perturbing anything else
multiplicatively, so a pre-1600 date can be replaced by an impossible one. Together
these affect 25 decided claims on Qwen2.5-7B, 3 of them parametric; excluding
all of them moves the rate from 8.2\% to 8.0\%. A parser over typed values would
remove all three.

\paragraph{Detection ceiling.} Section~\ref{sec:errors} locates the difficulty in claims
the model already knows, where AUROC falls to 0.651. We did not test whether a
signal distinguishing knowledge from use exists within black-box access, and we
regard that as the central open question this work leaves.

\section*{Ethical considerations}

This work studies a failure mode in which systems misrepresent the origin of
information they present. We do not release any model or method that increases the
incidence of that failure. Our detector, were it accurate enough to deploy, would
flag claims for human verification rather than suppress or rewrite them. We report
that it is not currently accurate enough for that use. The datasets contain no personal data beyond what is already
published in their sources. The blind validation reported in
Section~\ref{sec:validation} was carried out by three annotators, none of them an
author. They were graduate students at our institution who
volunteered and received no compensation. They were told the purpose of the task and
saw only questions, retrieved passages and model answers drawn from public
benchmarks.

\bibliography{custom}

@inproceedings{wallat2025correctness,
  title = {Correctness is not Faithfulness in Retrieval Augmented Generation Attributions},
  author = {Wallat, Jonas and Heuss, Maria and de Rijke, Maarten and Anand, Avishek},
  booktitle = {Proceedings of the 2025 International {ACM} {SIGIR} Conference on
               Innovative Concepts and Theories in Information Retrieval},
  pages = {22--32},
  publisher = {Association for Computing Machinery},
  doi = {10.1145/3731120.3744592},
  year = {2025}
}

@inproceedings{vandort2026howcite,
  title = {How Do {LLM}s Cite? {A} Mechanistic Interpretation of Attribution in {RAG}},
  author = {van Dort, Ian and Heuss, Maria},
  booktitle = {Advances in Information Retrieval},
  series = {Lecture Notes in Computer Science},
  volume = {16485},
  pages = {458--473},
  publisher = {Springer},
  doi = {10.1007/978-3-032-21324-2_35},
  year = {2026}
}

@inproceedings{worledge2024unifying,
  title = {Unifying Corroborative and Contributive Attributions in Large Language Models},
  author = {Worledge, Theodora and Shen, Judy Hanwen and Meister, Nicole and
            Winston, Caleb and Guestrin, Carlos},
  booktitle = {IEEE Conference on Secure and Trustworthy Machine Learning},
  year = {2024}
}

@article{rashkin2023measuring,
  title = {Measuring Attribution in Natural Language Generation Models},
  author = {Rashkin, Hannah and Nikolaev, Vitaly and Lamm, Matthew and
            Aroyo, Lora and Collins, Michael and Das, Dipanjan and
            Petrov, Slav and Tomar, Gaurav Singh and Turc, Iulia and Reitter, David},
  journal = {Computational Linguistics},
  volume = {49},
  number = {4},
  year = {2023}
}

@misc{bohnet2022attributed,
  title = {Attributed Question Answering: Evaluation and Modeling for Attributed
           Large Language Models},
  author = {Bohnet, Bernd and Tran, Vinh Q. and Verga, Pat and Aharoni, Roee and
            Andor, Daniel and Baldini Soares, Livio and Ciaramita, Massimiliano and
            Eisenstein, Jacob and Ganchev, Kuzman and Herzig, Jonathan and Hui, Kai and
            Kwiatkowski, Tom and Ma, Ji and Ni, Jianmo and Sestorain Saralegui, Lierni and
            Schuster, Tal and Cohen, William W. and Collins, Michael and Das, Dipanjan and
            Metzler, Donald and Petrov, Slav and Webster, Kellie},
  eprint = {2212.08037},
  archivePrefix = {arXiv},
  primaryClass = {cs.CL},
  year = {2022}
}

@inproceedings{gao2023enabling,
  title = {Enabling Large Language Models to Generate Text with Citations},
  author = {Gao, Tianyu and Yen, Howard and Yu, Jiatong and Chen, Danqi},
  booktitle = {Proceedings of the 2023 Conference on Empirical Methods in
               Natural Language Processing},
  pages = {6465--6488},
  year = {2023}
}

@inproceedings{liu2023evaluating,
  title = {Evaluating Verifiability in Generative Search Engines},
  author = {Liu, Nelson F. and Zhang, Tianyi and Liang, Percy},
  booktitle = {Findings of the Association for Computational Linguistics: EMNLP 2023},
  year = {2023}
}

@inproceedings{cohenwang2024contextcite,
  title = {{ContextCite}: Attributing Model Generation to Context},
  author = {Cohen-Wang, Benjamin and Shah, Harshay and Georgiev, Kristian and Madry, Aleksander},
  booktitle = {Advances in Neural Information Processing Systems},
  year = {2024}
}

@inproceedings{qi2024mirage,
  title = {Model Internals-based Answer Attribution for Trustworthy
           Retrieval-Augmented Generation},
  author = {Qi, Jirui and Sarti, Gabriele and Fern{\'a}ndez, Raquel and Bisazza, Arianna},
  booktitle = {Proceedings of the 2024 Conference on Empirical Methods in
               Natural Language Processing},
  pages = {6037--6053},
  year = {2024}
}

@inproceedings{li2026arcjsd,
  title = {Attributing Response to Context: A {Jensen-Shannon} Divergence Driven
           Mechanistic Study of Context Attribution in Retrieval-Augmented Generation},
  author = {Li, Ruizhe and Chen, Chen and Hu, Yuchen and Gao, Yanjun and
            Wang, Xi and Yilmaz, Emine},
  booktitle = {International Conference on Learning Representations},
  year = {2026}
}

@inproceedings{chuang2025selfcite,
  title = {{SelfCite}: Self-Supervised Alignment for Context Attribution in
           Large Language Models},
  author = {Chuang, Yung-Sung and Cohen-Wang, Benjamin and Shen, Zejiang and
            Wu, Zhaofeng and Xu, Hu and Lin, Xi Victoria and Glass, James R. and
            Li, Shang-Wen and Yih, Wen-Tau},
  booktitle = {Proceedings of the 42nd International Conference on Machine Learning},
  volume = {267},
  pages = {10839--10858},
  publisher = {PMLR},
  year = {2025}
}

@inproceedings{longpre2021entity,
  title = {Entity-Based Knowledge Conflicts in Question Answering},
  author = {Longpre, Shayne and Perisetla, Kartik and Chen, Anthony and
            Ramesh, Nikhil and DuBois, Chris and Singh, Sameer},
  booktitle = {Proceedings of the 2021 Conference on Empirical Methods in
               Natural Language Processing},
  pages = {7052--7063},
  year = {2021}
}

@inproceedings{xie2024adaptive,
  title = {Adaptive Chameleon or Stubborn Sloth: Revealing the Behavior of Large
           Language Models in Knowledge Conflicts},
  author = {Xie, Jian and Zhang, Kai and Chen, Jiangjie and Lou, Renze and Su, Yu},
  booktitle = {International Conference on Learning Representations},
  year = {2024}
}

@inproceedings{consens2025,
  title = {{ConSens}: Assessing Context Grounding in Open-Book Question Answering},
  author = {Vankov, Ivan and Ivanov, Matyo and Correia, Adriana and Botev, Victor},
  booktitle = {Artificial Neural Networks and Machine Learning, ICANN 2025},
  series = {Lecture Notes in Computer Science},
  volume = {16070},
  pages = {151--163},
  publisher = {Springer},
  doi = {10.1007/978-3-032-04549-2_13},
  year = {2025}
}

@misc{xu2026trustmargin,
  title = {{TrustMargin}: Training-Free Arbitration between Parametric Memory and
           Retrieved Evidence in Large Language Models},
  author = {Xu, Jingyan and Shi, Hong and Shan, Yi and Liu, Penghui and
            Bai, Yunhao and Li, Ningyuan and Liu, Xueyang},
  eprint = {2606.08397},
  archivePrefix = {arXiv},
  primaryClass = {cs.CL},
  year = {2026}
}

@inproceedings{maheshwari2025citefix,
  title = {{CiteFix}: Enhancing {RAG} Accuracy Through Post-Processing Citation Correction},
  author = {Maheshwari, Harsh and Tenneti, Srikanth and Nakkiran, Alwarappan},
  booktitle = {Proceedings of the 63rd Annual Meeting of the Association for
               Computational Linguistics: Industry Track},
  year = {2025}
}

@inproceedings{li2024citation,
  title = {Citation-Enhanced Generation for {LLM}-based Chatbots},
  author = {Li, Weitao and Li, Junkai and Ma, Weizhi and Liu, Yang},
  booktitle = {Proceedings of the 62nd Annual Meeting of the Association for
               Computational Linguistics},
  year = {2024}
}

@inproceedings{stelmakh2022asqa,
  title = {{ASQA}: Factoid Questions Meet Long-Form Answers},
  author = {Stelmakh, Ivan and Luan, Yi and Dhingra, Bhuwan and Chang, Ming-Wei},
  booktitle = {Proceedings of the 2022 Conference on Empirical Methods in
               Natural Language Processing},
  year = {2022}
}

@article{wilson1927probable,
  title = {Probable Inference, the Law of Succession, and Statistical Inference},
  author = {Wilson, Edwin B.},
  journal = {Journal of the American Statistical Association},
  volume = {22},
  number = {158},
  pages = {209--212},
  year = {1927}
}

@inproceedings{lewis2020retrieval,
  title = {Retrieval-Augmented Generation for Knowledge-Intensive {NLP} Tasks},
  author = {Lewis, Patrick and Perez, Ethan and Piktus, Aleksandra and Petroni, Fabio
            and Karpukhin, Vladimir and Goyal, Naman and K{\"u}ttler, Heinrich and
            Lewis, Mike and Yih, Wen-tau and Rockt{\"a}schel, Tim and Riedel, Sebastian
            and Kiela, Douwe},
  booktitle = {Advances in Neural Information Processing Systems},
  year = {2020}
}

@inproceedings{karpukhin2020dense,
  title = {Dense Passage Retrieval for Open-Domain Question Answering},
  author = {Karpukhin, Vladimir and O{\u{g}}uz, Barlas and Min, Sewon and Lewis, Patrick
            and Wu, Ledell and Edunov, Sergey and Chen, Danqi and Yih, Wen-tau},
  booktitle = {Proceedings of the 2020 Conference on Empirical Methods in
               Natural Language Processing},
  year = {2020}
}

@inproceedings{ni2022large,
  title = {Large Dual Encoders Are Generalizable Retrievers},
  author = {Ni, Jianmo and Qu, Chen and Lu, Jing and Dai, Zhuyun and
            {\'A}brego, Gustavo Hern{\'a}ndez and Ma, Ji and Zhao, Vincent Y. and
            Luan, Yi and Hall, Keith B. and Chang, Ming-Wei and Yang, Yinfei},
  booktitle = {Proceedings of the 2022 Conference on Empirical Methods in
               Natural Language Processing},
  year = {2022}
}

@inproceedings{fan2019eli5,
  title = {{ELI5}: Long Form Question Answering},
  author = {Fan, Angela and Jernite, Yacine and Perez, Ethan and Grangier, David
            and Weston, Jason and Auli, Michael},
  booktitle = {Proceedings of the 57th Annual Meeting of the Association for
               Computational Linguistics},
  year = {2019}
}

@inproceedings{amouyal2023qampari,
  title = {{QAMPARI}: A Benchmark for Open-domain Questions with Many Answers},
  author = {Amouyal, Samuel and Wolfson, Tomer and Rubin, Ohad and Yoran, Ori and
            Herzig, Jonathan and Berant, Jonathan},
  booktitle = {Proceedings of the Third Workshop on Natural Language Generation,
               Evaluation, and Metrics (GEM)},
  pages = {97--110},
  publisher = {Association for Computational Linguistics},
  address = {Singapore},
  year = {2023}
}

@inproceedings{he2021deberta,
  title = {{DeBERTa}: Decoding-enhanced {BERT} with Disentangled Attention},
  author = {He, Pengcheng and Liu, Xiaodong and Gao, Jianfeng and Chen, Weizhu},
  booktitle = {International Conference on Learning Representations},
  year = {2021}
}

@inproceedings{gao2023rarr,
  title = {{RARR}: Researching and Revising What Language Models Say, Using
           Language Models},
  author = {Gao, Luyu and Dai, Zhuyun and Pasupat, Panupong and Chen, Anthony and
            Chaganty, Arun Tejasvi and Fan, Yicheng and Zhao, Vincent Y. and
            Lao, Ni and Lee, Hongrae and Juan, Da-Cheng and Guu, Kelvin},
  booktitle = {Proceedings of the 61st Annual Meeting of the Association for
               Computational Linguistics},
  year = {2023}
}

@inproceedings{asai2024selfrag,
  title = {Self-{RAG}: Learning to Retrieve, Generate, and Critique through
           Self-Reflection},
  author = {Asai, Akari and Wu, Zeqiu and Wang, Yizhong and Sil, Avirup and
            Hajishirzi, Hannaneh},
  booktitle = {International Conference on Learning Representations},
  year = {2024}
}

@inproceedings{song2025measuring,
  title = {Measuring and Enhancing Trustworthiness of {LLM}s in {RAG} through
           Grounded Attributions and Learning to Refuse},
  author = {Song, Maojia and Sim, Shang Hong and Bhardwaj, Rishabh and
            Chieu, Hai Leong and Majumder, Navonil and Poria, Soujanya},
  booktitle = {International Conference on Learning Representations},
  year = {2025}
}

@inproceedings{yue2023automatic,
  title = {Automatic Evaluation of Attribution by Large Language Models},
  author = {Yue, Xiang and Wang, Boshi and Chen, Ziru and Zhang, Kai and
            Su, Yu and Sun, Huan},
  booktitle = {Findings of the Association for Computational Linguistics: EMNLP 2023},
  year = {2023}
}

@inproceedings{min2023factscore,
  title = {{FActScore}: Fine-grained Atomic Evaluation of Factual Precision in
           Long Form Text Generation},
  author = {Min, Sewon and Krishna, Kalpesh and Lyu, Xinxi and Lewis, Mike and
            Yih, Wen-tau and Koh, Pang Wei and Iyyer, Mohit and Zettlemoyer, Luke
            and Hajishirzi, Hannaneh},
  booktitle = {Proceedings of the 2023 Conference on Empirical Methods in
               Natural Language Processing},
  year = {2023}
}

@inproceedings{niu2024ragtruth,
  title = {{RAGTruth}: A Hallucination Corpus for Developing Trustworthy
           Retrieval-Augmented Language Models},
  author = {Niu, Cheng and Wu, Yuanhao and Zhu, Juno and Xu, Siliang and
            Shum, Kashun and Zhong, Randy and Song, Juntong and Zhang, Tong},
  booktitle = {Proceedings of the 62nd Annual Meeting of the Association for
               Computational Linguistics},
  year = {2024}
}

@inproceedings{wu2024clasheval,
  title = {{ClashEval}: Quantifying the Tug-of-War between an {LLM}'s Internal
           Prior and External Evidence},
  author = {Wu, Kevin and Wu, Eric and Zou, James},
  booktitle = {Advances in Neural Information Processing Systems Datasets and
               Benchmarks Track},
  year = {2024}
}

@inproceedings{su2024conflictbank,
  title = {{ConflictBank}: A Benchmark for Evaluating the Influence of Knowledge
           Conflicts in {LLM}s},
  author = {Su, Zhaochen and Zhang, Jun and Qu, Xiaoye and Zhu, Tong and
            Li, Yanshu and Sun, Jiashuo and Li, Juntao and Zhang, Min and Cheng, Yu},
  booktitle = {Advances in Neural Information Processing Systems Datasets and
               Benchmarks Track},
  year = {2024}
}

@inproceedings{ming2025faitheval,
  title = {{FaithEval}: Can Your Language Model Stay Faithful to Context, Even If
           ``The Moon is Made of Marshmallows''},
  author = {Ming, Yifei and Purushwalkam, Senthil and Pandit, Shrey and
            Ke, Zixuan and Nguyen, Xuan-Phi and Xiong, Caiming and Joty, Shafiq},
  booktitle = {International Conference on Learning Representations},
  year = {2025}
}

@inproceedings{xu2024knowledge,
  title = {Knowledge Conflicts for {LLM}s: A Survey},
  author = {Xu, Rongwu and Qi, Zehan and Guo, Zhijiang and Wang, Cunxiang and
            Wang, Hongru and Zhang, Yue and Xu, Wei},
  booktitle = {Proceedings of the 2024 Conference on Empirical Methods in
               Natural Language Processing},
  year = {2024}
}

@inproceedings{zhou2023context,
  title = {Context-faithful Prompting for Large Language Models},
  author = {Zhou, Wenxuan and Zhang, Sheng and Poon, Hoifung and Chen, Muhao},
  booktitle = {Findings of the Association for Computational Linguistics: EMNLP 2023},
  year = {2023}
}

@inproceedings{shi2024trusting,
  title = {Trusting Your Evidence: Hallucinate Less with Context-aware Decoding},
  author = {Shi, Weijia and Han, Xiaochuang and Lewis, Mike and Tsvetkov, Yulia
            and Zettlemoyer, Luke and Yih, Scott Wen-tau},
  booktitle = {Proceedings of the 2024 Conference of the North American Chapter of
               the Association for Computational Linguistics},
  year = {2024}
}

@inproceedings{monea2024glitch,
  title = {A Glitch in the Matrix? {L}ocating and Detecting Language Model Grounding
           with Fakepedia},
  author = {Monea, Giovanni and Peyrard, Maxime and Josifoski, Martin and
            Chaudhary, Vishrav and Eisner, Jason and Kiciman, Emre and
            Palangi, Hamid and Patra, Barun and West, Robert},
  booktitle = {Proceedings of the 62nd Annual Meeting of the Association for
               Computational Linguistics},
  year = {2024}
}

@inproceedings{wu2025retrieval,
  title = {Retrieval Head Mechanistically Explains Long-Context Factuality},
  author = {Wu, Wenhao and Wang, Yizhong and Xiao, Guangxuan and Peng, Hao and Fu, Yao},
  booktitle = {International Conference on Learning Representations},
  year = {2025}
}

@inproceedings{neeman2023disentqa,
  title = {{DisentQA}: Disentangling Parametric and Contextual Knowledge with
           Counterfactual Question Answering},
  author = {Neeman, Ella and Aharoni, Roee and Honovich, Or and Choshen, Leshem
            and Szpektor, Idan and Abend, Omri},
  booktitle = {Proceedings of the 61st Annual Meeting of the Association for
               Computational Linguistics},
  year = {2023}
}

@inproceedings{mallen2023when,
  title = {When Not to Trust Language Models: Investigating Effectiveness of
           Parametric and Non-Parametric Memories},
  author = {Mallen, Alex and Asai, Akari and Zhong, Victor and Das, Rajarshi and
            Khashabi, Daniel and Hajishirzi, Hannaneh},
  booktitle = {Proceedings of the 61st Annual Meeting of the Association for
               Computational Linguistics},
  year = {2023}
}

@inproceedings{slobodkin2024attribute,
  title = {Attribute First, then Generate: Locally-attributable Grounded Text Generation},
  author = {Slobodkin, Aviv and Hirsch, Eran and Cattan, Arie and Schuster, Tal
            and Dagan, Ido},
  booktitle = {Proceedings of the 62nd Annual Meeting of the Association for
               Computational Linguistics},
  year = {2024}
}

@inproceedings{ye2024effective,
  title = {Effective Large Language Model Adaptation for Improved Grounding and
           Citation Generation},
  author = {Ye, Xi and Sun, Ruoxi and Ar{\i}k, Sercan {\"O}. and Pfister, Tomas},
  booktitle = {Proceedings of the 2024 Conference of the North American Chapter of
               the Association for Computational Linguistics},
  year = {2024}
}

@inproceedings{sarti2024quantifying,
  title = {Quantifying the Plausibility of Context Reliance in Neural Machine
           Translation},
  author = {Sarti, Gabriele and Chrupa{\l}a, Grzegorz and Nissim, Malvina and
            Bisazza, Arianna},
  booktitle = {International Conference on Learning Representations},
  year = {2024}
}

@inproceedings{liu2025attribot,
  title = {{AttriBoT}: A Bag of Tricks for Efficiently Approximating Leave-One-Out
           Context Attribution},
  author = {Liu, Fengyuan and Kandpal, Nikhil and Raffel, Colin},
  booktitle = {International Conference on Learning Representations},
  year = {2025}
}

@inproceedings{jacovi2020towards,
  title = {Towards Faithfully Interpretable {NLP} Systems: How Should We Define and
           Evaluate Faithfulness?},
  author = {Jacovi, Alon and Goldberg, Yoav},
  booktitle = {Proceedings of the 58th Annual Meeting of the Association for
               Computational Linguistics},
  year = {2020}
}

@inproceedings{turpin2023language,
  title = {Language Models Don't Always Say What They Think: Unfaithful Explanations
           in Chain-of-Thought Prompting},
  author = {Turpin, Miles and Michael, Julian and Perez, Ethan and Bowman, Samuel R.},
  booktitle = {Advances in Neural Information Processing Systems},
  year = {2023}
}

@inproceedings{manakul2023selfcheckgpt,
  title = {{SelfCheckGPT}: Zero-Resource Black-Box Hallucination Detection for
           Generative Large Language Models},
  author = {Manakul, Potsawee and Liusie, Adian and Gales, Mark J. F.},
  booktitle = {Proceedings of the 2023 Conference on Empirical Methods in
               Natural Language Processing},
  year = {2023}
}

@book{efron1993introduction,
  title = {An Introduction to the Bootstrap},
  author = {Efron, Bradley and Tibshirani, Robert J.},
  publisher = {Chapman and Hall/CRC},
  year = {1993}
}

@article{cohen1960coefficient,
  title = {A Coefficient of Agreement for Nominal Scales},
  author = {Cohen, Jacob},
  journal = {Educational and Psychological Measurement},
  volume = {20},
  number = {1},
  pages = {37--46},
  year = {1960}
}

@article{qwen2025qwen25,
  title = {{Qwen2.5} Technical Report},
  author = {{Qwen Team}},
  journal = {arXiv preprint arXiv:2412.15115},
  year = {2025}
}

@article{grattafiori2024llama,
  title = {The {Llama} 3 Herd of Models},
  author = {Grattafiori, Aaron and others},
  journal = {arXiv preprint arXiv:2407.21783},
  year = {2024}
}

@inproceedings{yuan2026benchmarking,
  title = {Benchmarking {LLM}'s Capability in Reasoning over Conflicting Web References},
  author = {Yuan, Yizhen and Kong, Rui and Li, Dongze and Li, Yuanchun and Liu, Yunxin},
  booktitle = {Proceedings of the 64th Annual Meeting of the Association for
               Computational Linguistics (Volume 1: Long Papers)},
  pages = {303--322},
  address = {San Diego, California, United States},
  publisher = {Association for Computational Linguistics},
  doi = {10.18653/v1/2026.acl-long.11},
  year = {2026}
}

\clearpage
\appendix

\section{Why flagging cannot pay}
\label{sec:precision}

\begin{figure}[h]
\centering
\includegraphics[width=\columnwidth]{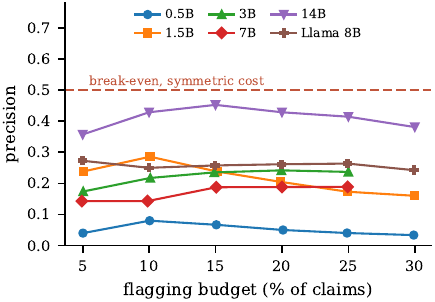}
\caption{Precision at the head of the ranking against flagging budget. Every model
stays below the break-even threshold at every budget, so each flag costs more
honesty than it buys.}
\label{fig:precision}
\end{figure}

Table~\ref{tab:repair} reports that flagging yields no honesty gain under symmetric
costs. Figure~\ref{fig:precision} shows why. Flagging a claim gains honesty when the claim is genuinely parametric and loses it
otherwise. Section~\ref{sec:repair} gives the expected gain at budget $b$ for a cost ratio
$c$; flagging is worthwhile only where precision exceeds $1/(1+c)$. Under symmetric costs
that threshold is 0.5.

No model reaches it at any budget. Qwen2.5-14B comes closest at 0.45, which follows from its higher base rate rather
than from a better detector. With one claim in five parametric, a ranker of the same
quality places more true positives at the head of the list. The gap between ranking quality and precision at the operating point is
the reason AUROC in the 0.70 to 0.88 range does not translate into a usable
flagging policy.

\section{Comparison with previously reported rates}
\label{sec:reconcile}

\begin{table}[h]
\centering\small
\begin{tabular}{lrr}
\toprule
Condition & Before & After \\
\midrule
ASQA Qwen2.5-1.5B & 12.4\% & 6.5\% \\
ASQA Qwen2.5-7B & 18.7\% & 8.2\% \\
ASQA Llama-3.1-8B & 22.3\% & 15.7\% \\
ELI5 Qwen2.5-7B & 14.7\% & 4.5\% \\
QAMPARI Qwen2.5-7B & 16.2\% & 2.4\% \\
\bottomrule
\end{tabular}
\caption{Measured rates before and after correcting the artefact, for the five
conditions measured under both protocols. Qwen2.5-0.5B, 3B and 14B were run after
the correction and have no uncorrected value.}
\label{tab:artefact}
\end{table}

\citet{wallat2025correctness} report post-rationalisation in 12\% to 57\% of
citations, against our 1.2\% to 15.7\%. Three differences account for the gap.

First, the protocols differ. They inject a statement the model has already written
into a passage that provides no evidence for it, and ask whether the model cites
that passage. We edit a value the model reported and ask which value it repeats.
Their test detects a citation attached without support; ours detects an answer
produced without the context.

Second, the upper end of their range comes from an adversarial condition designed to
elicit the behaviour rather than to estimate its base rate.

Third, and most consequentially, Table~\ref{tab:artefact} shows our own uncorrected
protocol reported 12\% to 22\% for the same reason described in
Section~\ref{sec:artefact}: a value left
visible in an unedited passage title. The uncorrected rates clustered between 12\% and
22\%, which we first read as a stable cross-dataset finding; the clustering was an
artefact of the leak. Rates measured under entity substitution are
sensitive to leakage of this kind, and we would expect any protocol in this family
to overestimate unless every surface carrying the value is edited. We report our
uncorrected numbers alongside the corrected ones so the size of that sensitivity is
visible.

\section{Entailment models}
\label{sec:nlisweep}

The support-based baseline in the body uses one cross-encoder. Concluding that support-based checking is uninformative
from a single checkpoint would not be a fair test. We repeat it with four: two sizes
of the same family, one with a different architecture, and one trained on more
inference data.

\begin{table}[h]
\centering\footnotesize
\setlength{\tabcolsep}{2.5pt}
\begin{tabular}{@{}lrrrrrr@{}}
\toprule
Verifier & 0.5B & 1.5B & 3B & 7B & 14B & L-8B \\
\midrule
DeBERTa-v3-base & 0.553 & 0.452 & 0.505 & 0.526 & 0.506 & 0.539 \\
DeBERTa-v3-large & 0.650 & 0.415 & 0.476 & 0.537 & 0.548 & 0.487 \\
RoBERTa-base & 0.530 & 0.538 & 0.547 & 0.573 & 0.515 & 0.543 \\
DeBERTa MNLI-F & 0.576 & 0.579 & 0.539 & 0.519 & 0.566 & 0.540 \\
\bottomrule
\end{tabular}
\caption{Support-based verification across four entailment models, AUROC, oriented
so that low support predicts parametric provenance. L-8B is Llama-3.1-8B.}
\label{tab:nlisweep}
\end{table}

Table~\ref{tab:nlisweep} gives the results. Across 24 measurements the values span
0.415 to 0.650 with a mean of 0.532;
excluding Qwen2.5-0.5B, whose six positive cases make its estimates unstable, the
range narrows to 0.415 to 0.579. No verifier separates provenance at a level that would be
useful. The highest value, 0.650, comes from the model with six positive cases and
carries an approximate interval of [0.41, 0.89], which contains chance. The failure is common to every checkpoint we tried, which is what the
structural argument predicts: entailment is satisfied by post-rationalised citations
because the model selects a passage that agrees with what it has written.

\section{Scanning candidate spans}
\label{sec:candidates}

The signal of Section~\ref{sec:precedes} is scored at a position our protocol chose.
A deployed detector receives an answer and does not know which span, if any, came
from memory, so it must score every numeric span and take the most suspicious.

The cost is unchanged. One forward pass yields logits at every position, so the
candidates are scored together rather than one at a time: two passes per answer, one
with the context and one without, exactly as before.

Candidates are the year and number shapes the protocol can substitute, excluding
values that also appear in the question, since a model can read those there whatever
the passages say. On Qwen2.5-7B this gives 3.8 candidates per answer.

The oracle figure here is 0.760 rather than the 0.776 of Table~\ref{tab:signals}
because the two score the same token under different continuations. A span's
log-probability is conditioned on everything preceding it, so scoring inside the
whole answer differs from scoring inside the sentence alone. Oracle and scan must
share a continuation to be comparable, which moves the oracle as well.

Taking the maximum change in entropy over candidates gives AUROC 0.668, against
0.760 when the span is known, a loss of 0.092 (paired bootstrap, $p = 0.002$). The
change in log-probability behaves similarly, 0.688 against 0.749, though that
difference is not significant ($p = 0.064$). The scan ranks the disputed span first
in 51.3\% of answers, about twice the 26.3\% expected by chance at this candidate
count.

Localisation therefore survives candidate selection at a measurable price. The
signal is weaker without the oracle, but it is not the oracle that creates it.

\section{Ablations}
\label{sec:ablations}

Three parameters of the protocol were varied to test whether the measured rates
depend on choices we made rather than on the models.

\paragraph{How far the value is moved.} Table~\ref{tab:ablation} shows the rate rises
by a factor of 3.6 on Qwen2.5-7B and 2.1 on Qwen2.5-14B between the near and far
settings. A near substitution is easier to accept, so the model follows the passages
more often; a distant one is easier to reject, so it falls back on memory. The
figures in Table~\ref{tab:rates} should therefore be read as conditional on the
default setting rather than as a constant. The ordering is consistent in direction:
Qwen2.5-14B is higher at all three distances, though the intervals overlap in each
and the gap narrows as the substitution becomes more extreme.

Part of what we measure is how implausible the substituted value is, and a model that
rejects an impossible date is behaving well rather than badly. The quantity is therefore better described as resistance to
contradicting evidence than as provenance in the abstract. The two coincide only
where the substituted value is plausible enough that rejecting it is not simply good
judgement. The near setting, two to five years from the truth, is the closest approximation to
that condition, and gives 4.6\% on Qwen2.5-7B and 8.5\% on Qwen2.5-14B, the two
models we ran at every distance.

\paragraph{How many passages are retrieved.} With Qwen2.5-7B the rate is 9.1\%,
9.3\%, 8.2\% and 7.7\% at one, three, five and ten passages. All four intervals
overlap and the extremes do not differ ($p = 0.47$). Retrieval depth does not drive
the phenomenon.

\paragraph{How many demonstrations are given.} The rate rises from 5.4\% with no
demonstrations to 8.2\% with two and 12.6\% with four ($p = 0.0003$ between the
extremes). Demonstrations teach the citation format, and a model that cites more
readily has more citable claims to be wrong about. This is the one prompt choice that
materially affects the measurement, and anyone reproducing the protocol should hold
it fixed.

\paragraph{Conflict flagging is not an artefact of implausible values.} Under the
near setting no planted date is ever impossible, and Qwen2.5-14B still flags 43
cases, 7.9\% of its records, against 2 for Qwen2.5-7B. Flagging rises with the
distance of the substitution, from 43 to 140 to 193 cases, which is what detection
should look like as the manipulation becomes more evident.

\section{Conflict flagging and qualitative examples}
\label{sec:examples}

Our marker list was written after reading Qwen2.5-14B's outputs, so it could be
tuned to that model's phrasing. Independently of markers, 58\% of its undecided
cases contain both values against 16\% for Qwen2.5-7B.

Nor is the model merely rejecting impossible dates. Our substitution clamps years at
2029, so a recent value can become a future one. Among Qwen2.5-14B's flagged cases
20\% carry the clamped value against a 15.8\% base rate, within sampling noise
($p = 0.17$), and four fifths involve an ordinary year contradicted only by what the
model knows.

These count as contextual in Table~\ref{tab:rates}: the answer asserts the
substituted value, which is what our definition requires, and the objection is
additional behaviour rather than a different provenance. Excluding them would raise
Qwen2.5-14B from 13.3\% to 20.0\%. This connects to work on conflicting evidence in retrieval
\citep{yuan2026benchmarking,wu2024clasheval}.

A conflict-flagged case from Qwen2.5-14B (Section~\ref{sec:conflict}):

\begin{quote}\small
\emph{When did the last season of Jersey Shore air?} The passages say 2029. The model
writes: ``\dots aired on December 20, \textbf{2029} [1][2]. Note that the dates
mentioned here seem unusually high \dots\ this could be an error in the provided
information.''
\end{quote}

Cases drawn from Qwen2.5-7B on ASQA, illustrating the two provenance classes. In each case a value appearing in both the answer and the
passages was replaced throughout the passages with a false variant, and the model was
queried again.

\paragraph{Claims produced from memory.} The passages assert the false value and the
model reports the true one.

\begin{quote}\small
\emph{Who plays Snowman in Smokey and the Bandit?} The passages say 1952. The model
writes: ``In the films Smokey and the Bandit (\textbf{1977}) \dots the character was
played by Jerry Reed [2][4][5].''
\end{quote}

\begin{quote}\small
\emph{When did the Seljuk Empire begin and end?} The passages say 1866. The model
writes: ``The Seljuk Empire began in \textbf{1037} when Tughril Beg founded it
[1][2].''
\end{quote}

Each carries a citation to passages asserting something else. A support-based
verifier sees a claim and a passage about the same subject, and finds the citation
supported.

\paragraph{Claims produced from context.} The model follows the planted value, even
where the result is plainly false.

\begin{quote}\small
\emph{When was the first Apple iPhone made?} The passages say 1967. The model
writes: ``The first iPhone was made and released on June 29, \textbf{1967} [2][4].''
\end{quote}

Such answers are factually wrong and provenance-faithful, which is the behaviour the
protocol is designed to identify. A model that asserts a 1967 iPhone on the strength
of an edited passage is context-faithful and factually incorrect. The two properties
are distinct, and optimising for one does not deliver the other.

\section{Signals that did not separate provenance}
\label{sec:discarded}

Three further hypotheses were tested and discarded. Together they bound what
black-box access achieves on this task.

We expected that interventions redirecting context support towards a competing
value would separate provenance better than deletion does. They did not. The one informative family
that does not reuse the label-generating edit reached AUROC 0.656, against 0.663 and
0.672 for the two control families.

We expected that absolute likelihood conflates memorisation with generic value
frequency, since a common year scores highly whatever its provenance. Marginal
likelihood alone predicts provenance at AUROC 0.505, so the confound is not present
in this data, and correcting for it costs performance (0.803 to 0.727).

We expected that the rate could be recovered without labels by fitting a
two-component mixture to the score distribution. At the separation these signals
achieve the fit is unstable, overestimating the rate substantially, and even 150
labels leave 0.042 absolute error. Counterfactual measurement is necessary rather
than merely convenient.

\section{Cost per claim}
\label{sec:cost}

Table~\ref{tab:cost} compares the cost of scoring a single claim. Our detector needs two
forward passes and no generation, so a system can afford to score every claim it
emits.

\begin{table}[h]
\centering\footnotesize
\setlength{\tabcolsep}{3pt}
\begin{tabular}{@{}lr@{}}
\toprule
Method & Cost per claim \\
\midrule
Localised detector (ours) & 2 forward \\
MIRAGE \citep{qi2024mirage} & 2 fwd, 1 bwd \\
ARC-JSD \citep{li2026arcjsd} & $|C| + 1$ forward \\
ContextCite \citep{cohenwang2024contextcite} & 32 to 256 forward \\
Counterfactual labelling & 1 generation \\
\bottomrule
\end{tabular}
\caption{The detector is cheaper than every attribution method it is compared against, and roughly two orders of magnitude cheaper than the regeneration used to
produce its own ground truth.}
\label{tab:cost}
\end{table}

\section{Reproducibility}
\label{sec:repro}

Code, configurations and per-run outputs are available at
\url{https://github.com/TanzeelAbbas/rag-citation-repair}. Each run records its configuration hash, git
commit, package and CUDA versions, and GPU model, so any number in this paper can be
traced to the code and settings that produced it. The cost per claim in
Appendix~\ref{sec:cost} is two forward passes, so the experiments reproduce on a
single GPU.

All generation was carried out on an L40S with PyTorch 2.6, Transformers 5.16 and
CUDA 12.4, in bfloat16.

\end{document}